# Causal Independence for Knowledge Acquisition and Inference

**David Heckerman**
Microsoft Research Center and
Department of Computer Science, UCLA
One Microsoft Way, 9S/1024
Redmond, WA 98052-6399
<heckerma@microsoft.com>

## Abstract

I introduce a temporal belief-network representation of causal independence that a knowledge engineer can use to elicit probabilistic models. Like the current, atemporal belief-network representation of causal independence, the new representation makes knowledge acquisition tractable. Unlike the atemproal representation, however, the temporal representation can simplify inference, and does not require the use of unobservable variables. The representation is less general than is the atemporal representation, but appears to be useful for many practical applications.

## 1 INTRODUCTION

When modeling the real world, we often encounter situations in which multiple causes bear on a single effect. A typical interaction of this sort can be modeled with the belief network shown in Figure 1. In the figure, the variable $e$ represents an effect and the variables $c^1, \ldots, c^n$ represent $n$ causes of that effect. This representation is inadequate, because it fails to represent the independence of causal interactions—or *causal independence*—that so often applies in this situation. Consequently, the representation imposes intractable demands on both knowledge acquisition and inference.

To overcome this inadequacy, knowledge engineers have used belief networks of the form shown in Figure 2 to represent causal independence (Kim and Pearl, 1983; Henrion, 1987; Srinivas, 1992). As in Figure 1, the variables $c^1, \ldots, c^n$, and $e$ represent the causes and effect, respectively. In addition, the intermediate variables $i^1, \ldots, i^n$ represent the independent contributions of each cause on the effect. That is, the effect $e$ is some deterministic function of these intermediate variables. This belief network encodes causal independence via the absence of arcs between pairs of $i$ variables and via the absence of any arc between a $c$ and $i$ variable. As a result, this representation avoid-s one limitation of the representation of Figure 1. In particular, the representation requires only $\mathcal{O}(n)$ probability assessments.

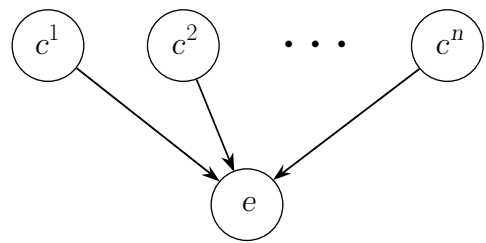

Figure 1: A belief network for multiple causes.

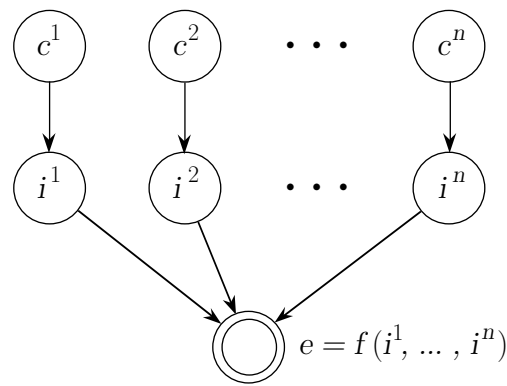

Figure 2: An explicit representation of causal independence.

Like the representation in Figure 1, however, this representation leads to intractable inference computations. In addition, the representation introduces a difficulty not present in the naive representation of multiple causes shown in Figure 1: The variables $i^1, \ldots, i^n$ are not observable. In my experience, assessments are easier to elicit (and presumably more reliable) when a person makes them in terms of observable variables.

In this paper, we examine a definition of causal independence that explicitly involves temporal considerations. From this definition, we derive a belief-network representation of causal independence. The representation facilitates tractable inference, and does not require the use of unobservable variables.

## 2 A TEMPORAL DEFINITION OF CAUSAL INDEPENDENCE

In this section, we examine a temporal definition of causal independence, which will form the basis of the belief-network representation of causal independence presented in the next section. In this definition, we associate a set of variables indexed by time with each cause and with the effect. We use $c_t^j$ to denote the variable associated with cause $c^j$ at time $t$, and $e_t$ to denote the variable associated with the effect at time $t$. For all times $t$ and $t'$, we require the variables $c_t^j$ and $c_{t'}^j$ to have the same set of instances. To simplify the definition, we assume that each variable has discrete instances. The generalization to continuous variables is straightforward. As will be clear from the discussion, there is no need to generalize to continuous time.

Under these assumptions, we can define causal independence to be the set of conditional-independent assertions

$$\forall t, c^j \;\; (e_{t+1} \perp c_t^1, \ldots, c_t^{j-1}, c_t^{j+1}, \ldots, c_t^n \mid \\ e_t, c_t^j, c_{t+1}^j, c_t^k = c_{t+1}^k \text{ for } k \neq j) \quad (1)$$

where $(X \perp Y|Z)$ denotes the conditional-independence assertion "the sets of variables $X$ and $Y$ are independent, given $Z$." Note that Assertion 1 is somewhat unusual, in that independence in conditioned, in part, on the knowledge that the instances of variables are equal, but otherwise undetermined ($c_t^k = c_{t+1}^k$ for $k \neq j$). In words, Assertion 1 states that if cause $c^j$ makes a transition from one instance to another between $t$ and $t+1$, and if no other causes makes a transition during this time interval, then the probability distribution over the effect at time $t+1$ depends only on the state of the effect at time $t$ and on the transition made by $c^j$; the distribution does not depend on the other causation variables.

## 3 A BELIEF-NETWORK REPRESENTATION OF CAUSAL INDEPENDENCE

As mentioned in the previous section, we can derive a belief-network representation of causal independence from this definition. First, for each cause, designate some instance of its associated variables to be *distinguished*. For most real-world models, this instance will represent the state of the cause in which that cause has no bearing on the effect—that is, the "off" state—but we do not require this association. Second, construct a belief network consisting of nodes $c^1, \ldots, c^n$, and $e_0, \ldots, e_n$, as shown in Figure 3. In this belief network, node $e_0$ represents the effect when all causes take on their distinguished instance. Node $c^1$ represents the state of cause $c^1$ after it has made a transition from its distinguished instance (a transition may be the trivial transition, wherein the cause maintains its distinguished instance). Node $e_1$ represents the effect after only $c^1$ has made the transition. In general, node $c^j$ represents the state of cause $c^j$ after it has made a (possibly trivial) transition from its distinguished instance. Node $e_j$ represents the effect after causes $c^1, \ldots, c^j$ have made their transitions. In particular, node $e_n$ represents the effect after all causes have made transitions. Thus, node $e_n$ corresponds to node $e$ in the atemporal representation of causal independence (Figure 2).

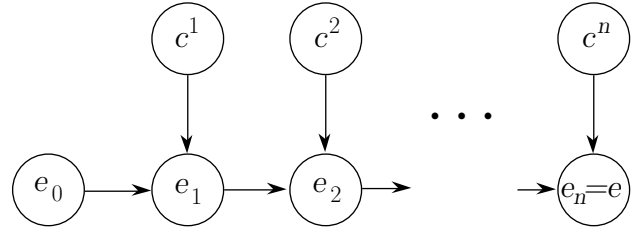

Figure 3: A temporal belief-network representation of causal independence.

The conditional independencies represented in the belief network of Figure 3 follow from the temporal definition of casual independence. The belief network, however, does not encode all of the conditional independencies associated with the definition. For example, from the temporal definition, we can obtain a belief network identical to the one in Figure 3, except for the exchange of nodes $c^j$ and $e_j$ with $c^k$ and $e_k$, respectively (for any $j \neq k$); we cannot obtain such a belief network directly from the belief network in Figure 3. Nonetheless, this belief-network representation of causal independence retains most of the advantages of the temporal definition. In particular, (1) probability assessment is tractable, (2) if we use any of the standard belief-network inference algorithms (e.g., Shachter (1986), Pearl (1985), or Lauritzen and Spiegelhalter (1988)), then inference is tractable, and (3) all probability assessments in this representation involve observable variables. The atemporal belief-network representation of causal independence does not have the latter two advantages.

Let us illustrate this representation with two examples. First, consider the most commonly used form of causal independence: the noisy OR-gate (Good, 1961; Suppes, 1970; Habbema, 1976; Kim and Pearl, 1983). This model, expressed in the atemporal representation of causal independence, consists of binary variables $c^1, \ldots, c^n$, $i^1, \ldots, i^n$, and $e$ (i.e., each variable has "true" and "false" as its only instances). Also, we require

$$f(i^1, \ldots, i^n) = i^1 \vee \ldots \vee i^n$$

and
$$p(i^j = \text{true}|c^j = \text{false}) = 0 \quad (2)$$
for $j = 1, \ldots, n$. The adjustable parameters of the model are the probabilities
$$p(i^j = \text{true}|c^j = \text{true}) \equiv q_j \quad (3)$$
Finally, to allow for the possibility that $e$ is true when all causes are absent—sometimes called a "leak" (Henrion, 1987)—we add a variable $c^0$ to the belief network in Figure 2, and instantiate $c^0$ to true.

The noisy OR-gate, expressed in the temporal belief-network representation consists of binary variables $c^1, \ldots, c^n$, and $e_0, \ldots, e_n$. The variable $e_n$ in this belief network corresponds to the variable $e$ in the atemporal representation. The distinguished instances of the variables $c^j$ are the instances "false." We require
$$p(e_j = \text{true}|e_{j-1} = \text{true}) = 1$$
which implements the OR function, and
$$p(e_j = \text{true}|e_{j-1} = \text{false}, c^j = \text{false}) = 0$$
which corresponds to Equation 2. The adjustable parameters of the model are the probabilities
$$p(e_j = \text{true}|e_{j-1} = \text{false}, c^j = \text{true}) = q_j$$
the same parameters as those in Equation 3. The probability $p(e_0 = \text{true})$ corresponds to the leak probability $q_0$ in the atemporal representation.

As mentioned, the advantages of the temporal representation are twofold. First, most inference computations—for example, the computation of $p(c_1|e = \text{true})$—using standard belief-network algorithms have time complexity $\mathcal{O}(n)$ in the temporal representation, but $\mathcal{O}(2^n)$ in the atemporal representation.[1] Second, the temporal representation requires probability assessments involving only observable variables. In contrast, the atemporal representation requires the assessments $p(i^j = \text{true}|c^j = \text{true})$, where the $i^j$ are unobservable variables.

Let us consider another model of a common cause-and-effect interaction: the *noisy adder*. This model, expressed in the atemporal representation of causal independence, consists of binary variables $c^1, \ldots, c^n$, and integer-valued variables $i^1, \ldots, i^n$. Each $i^j$ can take on values ranging from $-l$ to $+l$. Also, we require
$$f(i^1, \ldots, i^n) = i^1 + \ldots + i^n$$
and
$$p(i^j = 0|c^j = \text{false}) = 1 \quad (4)$$
for $i = 1, \ldots, n$. The adjustable parameters of the model are the probabilities
$$p(i^j = k|c^j = \text{true}) \equiv q_{jk} \quad (5)$$

---
[1] Pearl (1988) developed an $\mathcal{O}(n)$ inference algorithm for the noisy OR-gate interaction. The temporal representation eliminates the need for this special-case algorithm.

To allow for a leak, we add a variable $c^0$ to the belief network in Figure 2, and instantiate $c^0$ to true. Note that $e$ can take on values ranging from $-(n+1)l$ to $+(n+1)l$.

The noisy adder, expressed in the temporal representation consists of binary variables $c^1, \ldots, c^n$, and integer-valued variables $e_0, \ldots, e_n = e$. The distinguished instances of the variables $c^j$ are the instances "false." We require
$$p(e_j = k|e_{j-1} = k, c^j = \text{false}) = 1$$
which corresponds to the requirement of Equation 4. The adjustable parameters of the model are the probabilities
$$p(e_j = k|e_{j-1} = 0, c^j = \text{true}) = q_{jk}$$
the same parameters as those in Equation 5. We can derive the remaining probabilities from the additive model. In particular, we obtain
$$\forall m \quad p(e_j = k + m|e_{j-1} = m, c^j = \text{true}) = q_{jk}$$
Note that the values of variable $e_j$ range from $-(j+1)l$ to $+(j+1)l$. Consequently, inference using any standard belief-network algorithm has computational complexity $\mathcal{O}(n^3 l^2)$, a significant improvement over the intractable computations dictated by the atemporal representation.[2]

## 4 A REAL-WORLD EXAMPLE

Although the computational benefits of the temporal belief-network representation of causal independence are substantial, the new representation was inspired by the fact that it does not require probability assessments over unobservable variables. In particular, while I was developing a normative expert system for the morphologic diagnosis of blood disorders with the expert hematopathologist Patrick Ward, we encountered an interesting interaction between the possible diseases of the blood, a patient's white-blood-cell (W-BC) count, and various drugs that the patient may be taking as treatments for nonblood diseases. We attempted to model the interaction as a noisy adder, using the atemporal representation of causal independence, as shown in Figure 4(a). The pathologist did not understand clearly the definition of the intermediate variables, and could not provide the assessments required by the model. When I developed the alternative representation, and explained the probability assessments in terms of the belief network shown in Figure 4(b), the pathologist provided the assessments without difficulty.

---
[2] If we limit the integer values of $e$ to the range $[0, +l]$, then the computational complexity of inference becomes $\mathcal{O}(nl^2)$.

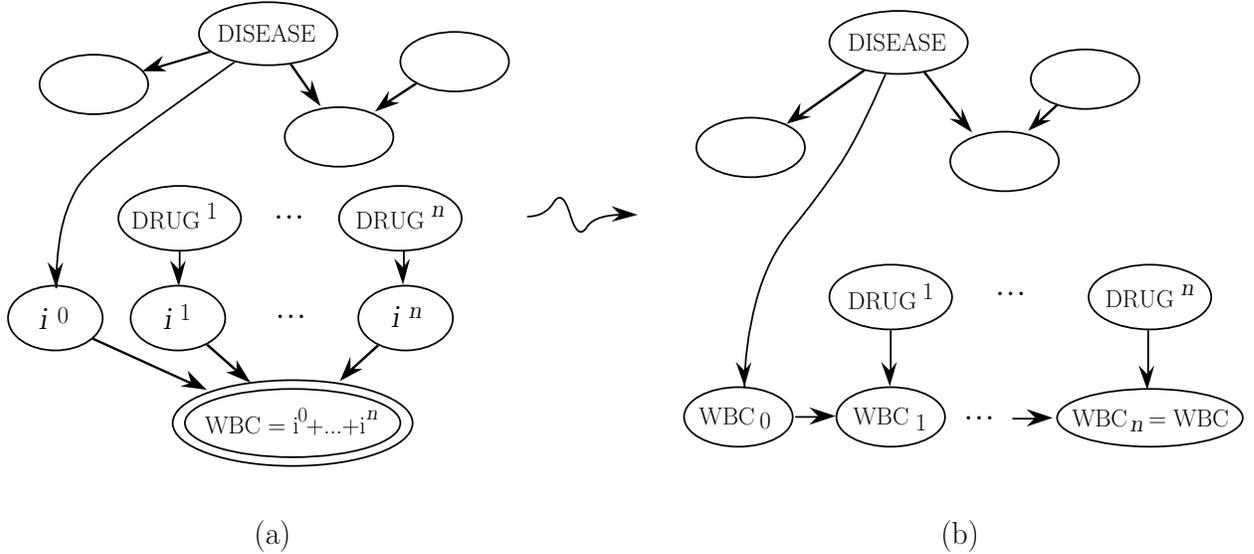

Figure 4: (a) A portion of a belief network for the morphologic diagnosis of blood disorders. The model uses an atemporal representation of causal independence. The variable DISEASE represents the mutually exclusive and exhaustive blood disorders. The variable WBC represents the patient's white-blood-cell count. The variables $DRUG^i$ represent various drug treatments for nonblood disorders. The variable $DRUG^0$, which is instantiated to true, is not shown. (b) A model of the same relationships using the temporal representation of causal independence.

## 5 IMPROVEMENTS IN THE REPRESENTATION FOR INFERENCE

As we discussed, the temporal belief-network representation does not represent all of the conditional independencies corresponding to the temporal definition of causal independence. In particular, the representation imposes a particular ordering over the causation variables. This order specification does not appear to impose limitations on knowledge acquisition, but situations may arise wherein this order specification may make inference inefficient. For example, suppose we know that the blood-disorder system described in the previous section is going to process a series of cases in which only the variables WBC, $DRUG^2$, and $DRUG^7$ will be instantiated. We would like the inference algorithm to recognize the irrelevance of the order of the causation variables, and rearrange the variables in the belief network so as to increase the efficiency of inference. In particular, the algorithm can rearrange the variables so that $DRUG^2$, $WBC^2$, $DRUG^7$, and $WBC^7 = WBC$ appear last in the chain. Consequently, for the entire series of cases, the algorithm need sum over the other drug and WBC variables only once.

To accomplish this goal, we can transform the belief network in Figure 4(b) to that in Figure 5. In particular, we reintroduce intermediate variables, and make the addition function explicit, using deterministic variables. This transformation can be done algorithmi-cally. Recognizing that addition is commutative, the inference algorithm now can rearrange $WBC^j$–$DRUG^j$ variable pairs so as to increase the efficiency of inference.

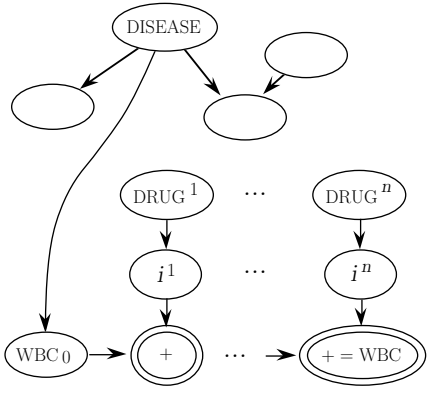

Figure 5: A modified version of the belief network in Figure 4. An inference algorithm may be able to use the additional information in this belief network to increase the efficiency of inference.

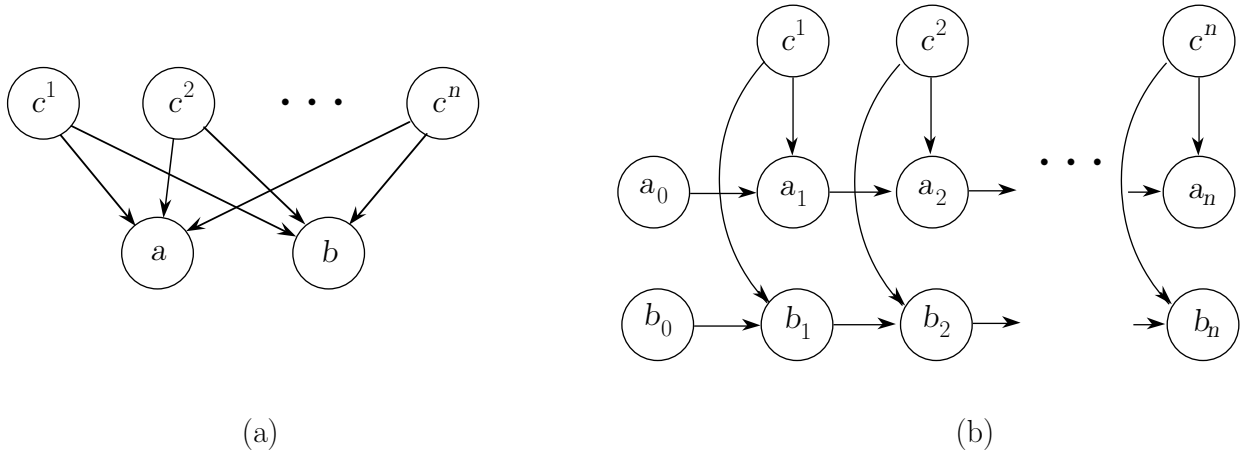

Figure 6: (a) Two effects share the same causes. (b) The same relations depicted in the temporal representation. Undirected cycles exist in both representations.

## 6 A LIMITATION OF THE REPRESENTATION FOR INFERENCE

Roughly speaking, two situations cause belief-network inference to become intractable: (1) a large parent set for one or more nodes, and (2) undirected cycles in the belief network. The use of the temporal representation for causal independence eliminates the first problem in many situations, but does not eliminate the second problem. To understand this point, consider the belief network shown in Figure 6(a), wherein $n$ causes bear on two effects $a$ and $b$. The transformation to the temporal representation of this situation, shown in Figure 6(b), does not eliminate undirected cycles. For example, Figure 6(b) contains the undirected cycle $b_1$—$c_1$—$a_1$—$a_2$—$c_2$—$b_2$—$b_1$. Consequently, the transformation does not produce tractable inference.

## 7 AN OBSERVATION ABOUT GENERALITY

The model in Figure 5 is a special case of the atemporal representation of causal independence, where the function $f$ is a nested series of added terms. That is,

$$f(\text{WBC}^0, i^1, \ldots, i^{n-1}, i^n) = \\ (((\ldots(\text{WBC}^0 + i^1) + \ldots) + i^{n-1}) + i^n)$$

Indeed, if we make the variables $i^j$ and function $f$ sufficiently complex, then any interaction encoded using the temporal belief-network representation of causal independence can be encoded using the atemporal representation of causal independence.

Despite its less general nature, however, the temporal representation appears to be useful for many practical applications. Over the last decade, I have participated in the construction of dozens of normative expert systems. I have reexamined the belief networks for these systems, and have found many instances in which the use of the temporal representation described in this paper would have simplified knowledge acquisition and inference. The added generality of the atemporal model would not have been useful in any of these instances.